\documentclass[unnumsec,webpdf,contemporary,large]{oup-authoring-template}
\usepackage{listings}

\usepackage[sectionbib,numbers,square]{natbib}

\graphicspath{{Fig/}}

\theoremstyle{thmstyleone}%
\theoremstyle{thmstyletwo}%
\theoremstyle{thmstylethree}%

\begin{document}
\bibliographystyle{plainnat}

\journaltitle{Preprint}
\copyrightyear{2023}
\pubyear{2023}

\firstpage{1}


\title[How to use Large Language Models]{How to use Large Language Models for Text Analysis}

\author[a,$\ast$]{Petter Törnberg}

\authormark{Petter Törnberg}

\address[a]{\orgdiv{Institute of Language, Logic and Computation (ILLC)}, \orgname{University of Amsterdam}
}

\corresp[$\ast$]{\href{email:p.tornberg@uva.nl}{p.tornberg@uva.nl}} \received{}{2023}{2023-07-25}

\abstract{This guide introduces Large Language Models (LLM) as a highly versatile text analysis method within the social sciences. As LLMs are easy-to-use, cheap, fast, and applicable on a broad range of text analysis tasks, ranging from text annotation and classification to sentiment analysis and critical discourse analysis, many scholars believe that LLMs will transform how we do text analysis. This how-to guide is aimed at students and researchers with limited programming experience, and offers a simple introduction to how LLMs can be used for text analysis in your own research project, as well as advice on best practices. We will go through each of the steps of analyzing textual data with LLMs using Python: installing the software, setting up the API, loading the data, developing an analysis prompt, analyzing the text, and validating the results. As an illustrative example, we will use the challenging task of identifying populism in political texts, and show how LLMs move beyond the existing state-of-the-art.}

\maketitle

Text analysis is a key task across social science disciplines -- from sociology, psychology, political science, to communication studies. As digitalization has meant that much of human communication is now stored and processed as digital data, the importance and potential of text analysis has exploded in recent years. At the same time, analyzing text remains challenging a challenging task.

While computational methods for analyzing textual data – such Natural Language Processing and Machine Learning – have evolved quickly in recent years, they are hard to use, often requiring deep knowledge in computational methods, as well as extensive manually coded training data. Even then, the methods often achieve only limited accuracy, as they struggle with sarcasm and irony, inferences that require contextual knowledge about the world, and key interpretive tasks such as putting oneself in the shoes of the text's author \cite{tornberg2021heterodox}. 

Humans have therefore been seen as the unrivaled gold standard for text analysis. At the same time, humans have important limitations of their own. Manually reading texts is slow and costly, limiting studies to relatively small samples – in particular for more in-depth interpretative tasks \citep{van2021validity}. As a result, manual text analysis has been criticized for bias, limited rigor and replicability, and low data quality \citep{tornberg2016combining}. 

The recent emergence of Large-Language Models (LLM), such as ChatGPT, may however be in the process of changing this situation -- and transforming how we do text analysis in the social sciences. ChatGPT is a pre-trained model based on a huge neural network with billions of parameters, trained on a substantial fraction of all the text on the Internet and in all books ever written \citep{brown2020language}. Such LLMs have demonstrated several surprising emergent capabilities, ranging from translation to programming \citep{wei_emergent_2022}. Researchers have found that LLMs are capable of carrying out nearly any text analysis task we throw at them. As these models are general rather than task-specific, they are even able to carry out tasks that traditional computational methods have failed at -- such as irony, sarcasm or subjective and contextual interpretation \citep{tornberg2023chatgpt}. Recent studies have found that LLMs perform well for a wide range of purposes, including ideological scaling \citep{wu_large_2023}, text annotation tasks \citep{gilardi_chatgpt_2023}, for simulating samples for survey research \citep{argyle_out_2022}, and much more \citep{bail2023can}.

As LLMs are easy-to-use, relatively cheap and fast, and applicable on a wide range of text analysis tasks, many scholars believe that LLMs represent a paradigm shift in text analysis in the social sciences. By enabling computational analysis of new tasks, they furthermore challenge the conventional division between the quantitative and qualitative realm. 

This how-to guide is aimed at students and researchers with limited knowledge in computational methods, who are interested in using LLMs in their own research projects. The guide offers a simple step-by-step introduction to using LLMs for text analysis.

As an illustrating example, we will draw on the important task of measuring populism in political texts. Within populism research, text analysis is seen as a direct way of measuring politicians' ideas as they are communicated to the public. However, measuring populism in text has proven a long-standing challenge. While manual content analysis has been considered the most accurate method, it is costly and does not scale well to large quantities of text. Dictionary-based approaches have been widely used, but studies show that they have limited validity \citep{rooduijn2011measuring}, as populist ideas tend to be latent and diffuse within a given text \citep{hawkins2018measuring}. Machine-learning techniques and dictionary approaches perform decently well within structured genres, such as party manifestos, within specific countries and time-periods \citep{hawkins2018textual}, but in cross-country, cross-context and cross-language texts, manual text analysis has thus far remained the only viable option for measuring populism \citep{hawkins2018measuring}. Scholars argue that measuring populism requires context-dependent understanding, knowledge of the country-government period, and the need to use longer passage of text as unit of measurement, rather than individual words \citep{hawkins2018measuring, manucci2017big, rooduijn2011measuring} -- skills that have thus far been outside the reach of computational methods. As we will see, LLMs have changed this situation, and thus potentially offers to solve one of the most central and basic issues within in the literature, by offering a valid, reproducible and precise way of measuring populism across languages, contexts and regions. 

The complete example code, including detailed descriptions, can be found at: \href{https://github.com/cssmodels/howtousellms}{github.com/cssmodels/howtousellms}.

\section{What are Large Language Models?}\label{whatarellms}

Large Language Models are advanced artificial intelligence (AI) systems designed to interpret and generate human language. These models employ deep learning techniques based on artificial neural networks -- essentially abstract mathematical models of brains -- and utilize vast amounts of textual data to learn patterns, semantics, and grammar. 

The currently most famous LLM is ChatGPT. ChatGPT is an AI chatbot developed by OpenAI and introduced in November 2022. ChatGPT simulates a conversation with the user. It is part of the Generative Pre-trained Transformer (GPT) family of language models and is based on OpenAI's LLMs GPT-3.5 and GPT-4 \citep{brown2020language}. The GPT models were trained on a truly vast corpus of text, and then fine-tuned to achieve human-like responses \citep{ziegler_fine-tuning_2019}, by having human trainers acting as both the user and the AI assistant in simulated conversations \citep{schulman_chatgpt_2022}.

While smaller versions of these transformer-based language models worked like a form of sophisticated autocomplete, the larger models began taking on surprising emergent properties -- even gaining capacities for which they were not explicitly trained. ChatGPT, for instance, seems to compose new sentences and information, rather than just regurgitating previous phrases. The chatbot can furthermore operate in and translate between several languages, write prose or poetry on any topic in a given style, and even generate programming code. 

For social scientists, one of the most important such emergent capacity is that the models can analyze nearly any textual statement. As researchers, we can ask the model nearly any question about a given text, including identifying themes or topics, whether the text expresses misinformation, what emotions are expressed in the text, or the possible intentions of the author \citep{tornberg2023chatgpt}. New tasks for which the models excel are still being discovered. Early studies have shown that the models can even outperform humans when it comes to interpretive textual analysis, showing higher accuracy, lower bias, and higher reliability across languages and regional contexts than human experts \citep{tornberg2023chatgpt}.

\section{When are LLMs an appropriate method?}\label{appropriate}

LLMs can thus carry out a range of text analysis tasks, including tasks that have previously been beyond the realm of computational methods, such as subjective interpretation, contextual inferences, and understanding sarcasm and irony. At the same time, some particular caveats must be made, not least since precise limitations and characteristics of the models are yet to be fully understood. 

First, LLMs can be sensitive to researcher choices, such as prompt design. Just as human coders, LLMs are prone to misunderstanding the task if it is not precisely specified. Formulating the exact instructions for the model should thus be understood as the key conceptual task within LLM-supported text analysis.

Second, LLMs embody the biases and prejudices of the texts on which they are trained. ChatGPT, for instance, has been found to display problematic gender and racial stereotypes. It is important to be aware that such biases may affect the models' results on text analysis tasks. The models should not be understood as a means of achieving ``objectivity'' or ``impartiality'', but must be treated in similar ways as data from human coders -- expressing certain perspectives and particular biases.

Third, since it is not known precisely how the models acquire their emergent capacities, we do not know their scope or limitations. It is therefore important to be cautious that they may fail in unexpected ways on specific tasks or challenges. 

LLMs are thus best thought of as a student assistant that can be instructed for textual analysis tasks: while highly versatile and capable to engage in a range of tasks, they are also prone to misunderstand our instructions, and may display certain biases. We must therefore use care and make sure to validate the results against other methods. 

Another limitation is speed and costs. While LLMs are significantly faster than any human classifier, they are much slower than most traditional natural language processing methods, such as machine learning classifiers. An LLM analysis request can take up to several seconds, depending on its complexity. Datasets that involve millions or tens of millions of datapoints can therefore be prohibitively time-consuming to analyze, instead requiring taking relevant samples. This situation may however change as the technology develops.

There are also limits on the length of texts that can be processed, due to the so-called ``context window'': the range of ``tokens'' that the model can consider when generating responses. A token corresponds to roughly 2/3 of a word. GPT-3 models have a window size of 2000 tokens, while GPT-4 has a window size of 32,000. While the context window is quickly increasing as the technology improves, it may be necessary to considered when we design our analysis. It may, for instance, be necessary to split texts into several smaller chunks to be analyzed part by part, and then reassembling the results in the post-analysis.

Finally, there are some important ethical considerations to take into account. First, in many cases, the platform running the LLM – OpenAI in the case of ChatGPT – will receive access to any data that you send to their API. The API should therefore never be used with sensitive or confidential material. Second, the limitations and potential biases of the method should be transparently reported and carefully considered, to make sure that the research does not inadvertently contribute to perpetuating existing biases. Finally, the usual ethical considerations for using digital data apply also when using LLMs, such as making sure that the privacy of users is maintained (see existing ethical guidelines for internet research, such as \citep{franzke2020association,british2017british}). As our methods before more powerful, we need to be even more thoughtful of the privacy implications and the use of public posts without explicit consent. The use of LLMs may grant access to deep insights about texts that their authors may not have expected when they made those texts public -- thus raising challenging questions.

We will now turn to a practical step-by-step guide to how you can use LLMs in your own text analysis projects.

\section{Step 1. Signing up for API access}
There are several LLMs, and new ones are frequently released. In this guide, we will use OpenAI's ChatGPT. However, it will be easy to adapt this guide to any LLM that offers an API.

To use the ChatGPT API, we must first request access from OpenAI. 

\begin{enumerate}
  \item Go to the OpenAI API website at \href{https://platform.openai.com/}{platform.openai.com}. Click ``sign up'' and follow the instructions to register. You will be guided through the API access signup process, which includes providing your contact information, providing payment details, and agreeing to the terms of service.
  \item 	On the API page, you will find details about the ChatGPT API, including pricing, documentation, and usage guides. Review this information to understand the API's capabilities and pricing structure.
  \item If your request is approved, you will receive a confirmation. To use the API, you need your API key from OpenAI. The key can be found by logging in to your account, and navigating to the page ``View API keys''. Store your key in a safe place.
\end{enumerate}

To use the OpenAI API from Python, you first need to set up Python and the relevant library. You are recommended to install Python and Jupyter Notebook, which offers an easy-to-use environment for coding, in which you can make use of the Notebook that is associated to this guide. Find and follow instructions for how to install Python and Jupyter Notebook on your particular system.

Install the \textit{openai} Python library: OpenAI provides an official Python library that simplifies API interactions. You can install it and import using pip by running the following command in a Jupyter Notebook.

\begin{minipage}{\hsize}
\tiny \begin{lstlisting}[language=Python,frame=single,framexleftmargin=10pt,framexrightmargin=-5pt,framesep=5pt,linewidth=0.98\textwidth]
!pip install openai
!pip install pandas
import pandas as pd
import openai 
openai.api_key = 'YOUR_API_KEY' 
\end{lstlisting}
\end{minipage}

The final line sets your API key that you received from the OpenAI website.

To interact with the ChatGPT API, you can now use the openai.Completion.create() method. The API will return a response object containing the generated output, and you can access the generated text using response.choices[0].text. For example, the following example asks ChatGPT what the meaning of life is, and writes the answer in the notebook.

The \textit{model} parameter specifies the model to use, the messages parameter describes an existing conversation between the chatbot and the user, and the \textit{max\_tokens} parameter determines the length of the generated response. The \textit{temperature} parameter controls the randomness or creativity of the model's output, where higher values (e.g., 0.8) make the output more diverse and unpredictable, and lower values (e.g., 0.2) make it more focused and deterministic. 

\begin{minipage}{\hsize}
\tiny \begin{lstlisting}[language=Python,frame=single,framexleftmargin=10pt,framexrightmargin=-5pt,framesep=5pt,linewidth=0.98\textwidth]
response = openai.ChatCompletion.create(
  model = 'gpt-4', 
  temperature=0.2,
  max_tokens=50,
  messages=[
    {"role": "user", 
    "content": 
    ``What is the meaning of life?''}]
)
result = ''
for choice in response.choices:
  result += choice.message.content
print(result)
#'As an AI, I don't have personal beliefs or experiences. 
# However, I can tell you that the meaning of life is 
[...]'
\end{lstlisting}
\end{minipage}

Be mindful that while using the ChatGPT API is cheap, it is not free and there may furthermore be usage limits. You should therefore be mindful of your calls, and you may  want to set a maximum expense cap on the OpenAI platform to avoid unexpectedly high costs.

\section{Step 2. Loading and preparing your data}\label{subsec2}
Before running the LLM, we need to load our text data into a processable format. Unlike traditional computational approaches, such as topic modeling or machine learning, there is no need to preprocess the text data with tokenization or lemmatization: the LLM can read unprocessed and messy data, and even handle text in different languages. 

The only necessary processing is to make sure that our texts are short enough to fit into the context window of our model, since LLMs cannot process texts that are longer. To do so, we need to either split our texts into several pieces, or truncate it to the maximal length, which will depend on the model that we use. In the case of our populist speeches, we use the 32K version of ChatGPT-4, which has a long enough context window to handle all the included speeches. 

For the purposes of this guide, we will load the data into a \textit{pandas dataframe}, to allow analyzing it using the LLM. Python \textit{pandas} can read a range of formats, and the details of loading the data depends on the details of your project. 

In our example, we use the database of speeches of the Global Populism Database \cite{hawkins2019global}. This prominent database offers a human coded levels of populism for a large number of speeches. The database provides a CSV file which lists the data points, and plain-text (.txt) files with the texts. We can easily load the text and metadata into a \textit{dataframe} using pandas \textit{pd.read\_csv(...)} command.

\section{Step 3. Prompt engineering}\label{prompt}
Prompt engineering refers to the formulation of instructions to LLMs for specific tasks or objectives: the instruction that tell the LLM how to analyze our text. The details of the prompts are important for guiding the model's behavior and influences the output. Just like guiding an assistant to carrying out a text analysis task, so prompt engineering is an important skill. 

The instruction to the model can be understood as the way in which a particular social scientific concept is encoded. Prompt engineering is in itself a form of qualitative method, in that one seeks to formulate an instruction that best captures some aspect of social reality. 

In formulating an initial prompt, several aspects should be taken into consideration:

\begin{enumerate}
  \item Define the task: Clearly define the objective of your text analysis task to be able to evaluate what the model produces. Determine what specific information or insights you aim to extract from the text data. 
  \item Determine the desired output: Identify the type of response or output you expect from the language model. Are you seeking factual information, subjective opinions, or predictive insights? 
  \item Consider length and specificity: Determine the appropriate length and level of specificity for your prompt. A concise and focused prompt may guide the model more effectively, while a broader prompt might encourage a more exploratory response. Strike a balance based on your analysis objectives.
  \item Include instructions or context: Incorporate relevant instructions or contextual cues in your prompt to guide the language model’s behavior. This could involve specifying the type of response desired, providing background information, or asking the model to consider specific aspects of the text.
  \item Start from existing research: A good starting point for a prompt is often to look at the instructions given to human coders to carry out similar coding tasks.
  \item Make results parseable: You will need to programmatically parse the output of the model. For this to be possible, the model has to respond in a consistent and systematic way. For instance, if you are seeking 4 numeric codes, you can append the following to your question: '[Answer in the format: ``0, 1, 2, 3''. Do not motivate your answer.]'
  \item Iterate and test: Prompt development often involves an iterative process. Experiment with different prompt formulations and test them with your text analysis task. Observe the outputs generated by the language model and refine the prompt accordingly to improve the results. (See below.)
\end{enumerate}

In our example, we are seeking to measure populism in political discourse. We follow the influential definition of populism suggested by Cas Mudde \citep{mudde2017populism}, which describes populism as ``an ideology that considers society to be ultimately separated into two homogeneous and antagonistic groups, 'the pure people' versus 'the corrupt elite', and which argues that politics should be an expression of the volonté générale (general will) of the people.'' As a starting point, we draw on existing human coder instructions. While the Global Populism Database \cite{hawkins2019global} unfortunately does not provide the instructions that their coders were given, we draw on an instruction file from the same research institution 
\citep{populismcoding}. Our initial instruction reads as follows:

\textit{instruction = ``Your task is to evaluate the level of populism in a political text. [...] A populist text is  characterized by BOTH of the following elements: 1. People-centrism: how much does 
the text focus on "the people" or "ordinary people" as an  indivisible or homogeneous community? Does the text  promote a politics as the popular will of "the people"? Appeals to specific subgroups of the population (such as ethnicities, regional groups, classes) are inherently antithetical to populism. 2. Anti-elitism: how much does the text focus on "the elite", and to what extent are elites in general described in negative terms? In populist texts, the elite is often described as corrupt, and the juxtaposition between the ordinary people and the elite is cast as a moral struggle between good and bad.  Criticism of specific elements within an elite is not populist: a populist appeal must regard the elite in its entirety as anathema. [...] [Answer with a number in the 0-2 range, followed by a semi-colon, and then a brief motivation. For instance: ``1.23; The text shows many elements of a populist text.'' Do not use quotation marks.]''}

Following our validation (see below), we can further improve our prompt depending on how well the LLM reproduces the intended results.
  
\section{Step 4. Calling the LLM and analyzing the results}

We have now set up the API access, loaded our data, and formulated our analysis prompt. We can now proceed to analyzing our data.

We first define a function that calls the API and handles any errors:

\begin{minipage}{\hsize}
\tiny \begin{lstlisting}[language=Python,frame=single,framexleftmargin=10pt,framexrightmargin=-5pt,framesep=5pt,linewidth=0.98\textwidth]
def analyze_message(text, instruction, 
 model = 'gpt-4', temperature=0.2):
  [...]
  response = openai.ChatCompletion.create(
  model = model, 
  temperature=temperature,
   messages=[
    {"role": "system", 
     "content": f"'{instruction}'"},
     {"role": "user", 
     "content": f"'{text}'"}])
# Handle errors
[...]
\end{lstlisting}
\end{minipage}

We can then define a function that parses the results from the LLM: it takes the text returned from the LLM and transform it into data we can use for our further analysis. The details of this function will depend on the call to the LLM. In our example, we asked for a number 0-2, followed by a semi-colon, followed by a motivation. We can therefore split on the semi-colon, and use the first part as the number, and the second part as the motivation:

\begin{minipage}{\hsize}
\tiny \begin{lstlisting}[language=Python,frame=single,framexleftmargin=10pt,framexrightmargin=-5pt,framesep=5pt,linewidth=0.98\textwidth]
def parse_result(result):
  return result.split(';', 2)
\end{lstlisting}
\end{minipage}

We can now finally create the main part of our code: a loop that goes through our data, analyzes the texts one by one, parses the answers from the API, and stores the result. If the process crashes -- for instance due to our laptop running out of battery -- the process will pick up where it was when we restart it.

\begin{minipage}{\hsize}
\tiny \begin{lstlisting}[language=Python,frame=single,framexleftmargin=10pt,framexrightmargin=-5pt,framesep=5pt,linewidth=0.98\textwidth]
df = pd.read_pickle(filename)
model = 'gpt-4'
i = 0	
while(True):
  left = df.loc[df['answer'].isna()]  
  if len(left)==0:
    print("Finished!")
    break
  line = left.sample()
  index = line.index.values[0]
  time.sleep(1)
  result = analyze_message( 
    line['text_truncated'].values[0], 
    instruction, model = model)
  
  #Parse the result
  answer,motivation = parse_result(result)
  
  #Store the result
  df.loc[index,'answer'] = answer
  df.loc[index,'motivation'] = motivation
  df.to_pickle(filename)
\end{lstlisting}
\end{minipage}

We now have our result in the dataframe and stored persistently in a file. We now need to validate, to see whether the model produces the intended result.

\section{4. Validation}
Careful validation is essential to make sure that the models are measuring what we intend -- and that they do so without problematic biases. To validate our models, we can compare the outputs with established benchmarks, ground truth data, or expert evaluations to validate the effectiveness in achieving the desired analysis outcomes.

For many tasks, a simple way of validating is to extract a sample of the data, and have human coders manually classifying the items. We can do so by simply extracting an Excel file that the coders can open in Excel. After they have finished coding, we load the Excel files back into Python, and use the results to validate the model. In our populist case, however, we instead use the existing populism database to validate our data. 

To measure the correspondence between the model result and our validation data, we can use the Krippendorff’s alpha \cite{de2012calculating}. The Krippendorff’s alpha gives a measure of interrater agreement, and is used to assess the extent to which multiple raters or coders agree when coding or categorizing qualitative data. Krippendorff’s alpha takes into account both the observed agreement among raters and the expected agreement by chance. It can be applied to different types of nominal, ordinal, or interval-level data. The calculation of Krippendorff's alpha considers the number of coders, the number of categories or codes, and the observed and expected agreement. It quantifies the agreement beyond what is expected by chance, taking into account the distribution of codes and the variation among raters. The coefficient ranges from 0 to 1, with higher values indicating greater agreement among coders.

In Python, the simpledorff library provides a simple way to measure the Krippendorff's alpha. We can apply this on the validation data using: 

\begin{minipage}{\hsize}
\tiny \begin{lstlisting}[language=Python,frame=single,framexleftmargin=10pt,framexrightmargin=-5pt,framesep=5pt,linewidth=0.98\textwidth]
!pip install simpledorff
import simpledorff

simpledorff.calculate_krippendorffs_alpha_for_df(
df, experiment_col=experiment_col,
annotator_col=annotator_col, class_col=class_col)
\end{lstlisting}
\end{minipage}

The resulting a Krippendorff's alpha of 0.635. This is a relatively high correspondence -- at least considering that the LLM was not even given the same coding instructions as the original coders. The human coders of the initial data have a Krippendorff's alpha (interval-level) of 0.827, however these coders worked in conjunctions, comparing their results and addressing disagreements, suggesting that we should expect significantly higher intercoder agreement. 

\subsection{\textbf{Iterative process of concept and prompt development}}
Our first result should however only be seen as our starting-point, rather than our final destination. When using LLMs for text analysis, it is recommended to use an iterative process of simultaneous prompt and concept development. 

A useful strategy is to engage in a closer examination of the cases where the LLM and the coders disagree, and compare the motivations provided by the model with those provided by the human coders. Such a process can also in itself be deeply informative about the data and concept. This should not be thought of as merely the fine-tuning of the LLM instructions to match the human data: as research has suggested, the LLM can in fact achieve higher accuracy than human coders, and the result of human coding hence cannot be treated as an unquestioned gold standard \citep{tornberg2023chatgpt}. Instead, it may be more productive to think of it a process of mutual learning through an iterative process for constructing rigorous and reproducible definitions of scientific concepts. 

In our populist example, we do not have the luxury of working with the original coders or accessing their motivations for their coding decisions. However, we can manually examine the texts for which the LLM and the coders disagree, and form our own judgements. To do so, we merge the human coded data and the LLM data, sort by disagreement, and save as a CSV to examine the data manually:

\begin{minipage}{\hsize}
\tiny \begin{lstlisting}[language=Python,frame=single,framexleftmargin=10pt,framexrightmargin=-5pt,framesep=5pt,linewidth=0.98\textwidth]
wrong = df2.merge(val, on=`merging_variable')
wrong['diff'] = abs(wrong['answer_x']-wrong['answer_y'])
wrong.sort_values(['diff']).to_csv('disagreements.csv') 
\end{lstlisting}
\end{minipage}

To take an example, one of the texts for which there is most disagreement is a speech by Italian Prime Minister Berlusconi. The speech is rated as populist (1.0) by the human coders, but as not populist (0.0) by the LLM, with the motivation: ''The text does not contain populist elements. It does not focus on `the people' as a homogeneous group, nor does it depict `the elite' as a corrupt entity. Instead, it focuses on historical events, the importance of freedom, and the unity of the nation.'' Examining the content of the speech (see the associated Notebook), the speech focuses on the Italian soldiers that fought against the Nazis during World War II, and calls for unity across ethnic and religious groups. While such calls for unity could perhaps be construed as promoting a notion of ``the people'', it seems far-fetched to view support for resistance against actual Nazis as a form of populism. Examining the speech, it thus seems the LLM may have been reasonable in classifying it as non-populist. 

When we have finished developing and validating our model, we can apply it to new datasets. For instance, our populism model can be used to develop the type of large international comparative database of populism that has thus far been limited by the need for manual coding.

\section{Conclusion}
LLMs are in the process of transforming text analysis in the social sciences, by offering a powerful and versatile method for examining large text data. This how-to guide has offered a practical step-by-step introduction to using LLMs in your own research project. 

While the guide has focused primarily on using LLMs for coding data for quantitative analysis, the method can be similarly employed for qualitative analysis. A similar approach as outlined in this guide can be used to employ LLMs to support a process of close-reading, by having it identify key texts or latent patterns in large textual datasets. LLMs challenge the conventional division between qualitative and quantitative methods in text analysis, and as the method is so new, its potential is still being uncovered. Your own study can contribute to this by finding imaginative ways of using LLMs to move beyond existing methodological limitations, and finding new ways of making sense of the social world. 

\section{Supplementary Material}
The Juputer Notebook that is associated to this guide can be found on: \href{https://github.com/cssmodels/howtousellms}{https://github.com/cssmodels/howtousellms}  

\bibliography{reference}

\end{document}